# Developing a Data-Driven Categorical Taxonomy of Emotional Expressions in Real World Human Robot Interactions

Ghazal Saheb Jam
School of Computing Science
Simon Fraser University
Burnaby, BC, Canada
gsahebja@sfu.ca

Jimin Rhim
School of Computing Science
Simon Fraser University
Burnaby, BC, Canada
jrhim@sfu.ca

Angelica Lim
School of Computing Science
Simon Fraser University
Burnaby, BC, Canada
angelica@sfu.ca

## ABSTRACT
Emotions are reactions that can be expressed through a variety of social signals. For example, anger can be expressed through a scowl, narrowed eyes, a long stare, or many other expressions. This complexity is problematic when attempting to recognize a human's expression in a human-robot interaction: categorical emotion models used in HRI typically use only a few prototypical classes, and do not cover the wide array of expressions in the wild. We propose a data-driven method towards increasing the number of known emotion classes present in human-robot interactions, to 28 classes or more. The method includes the use of automatic segmentation of video streams into short (<10s) videos, and annotation using the large set of widely-understood emojis as categories. In this work, we showcase our initial results using a large in-the-wild HRI dataset (UE-HRI), with 61 clips randomly sampled from the dataset, labeled with 28 different emojis. In particular, our results showed that the "skeptical" emoji was a common expression in our dataset, which is not often considered in typical emotion taxonomies. This is the first step in developing a rich taxonomy of emotional expressions that can be used in the future as labels for training machine learning models, towards more accurate perception of humans by robots.

## 1 INTRODUCTION

Nonverbal communication is essential for seamless human robot interaction (HRI). Emotional expressions in particular have been highly studied in HRI, and emotion taxonomies from the psychology domain are typically used to represent the nonverbal affective cues expressed by a human [1]. Generally, there are three main types of affect representation models: categorical, dimensional, and hybrid.

Categorical models include discrete states, each representing a particular affect category [41]. Initial work on categorical models of emotion focused primarily on six states. These states were happiness, surprise, disgust, anger, sadness and fear [16] [13]. Later, researchers added neutral to these categories [45] [20] and generated new categories by combining these basic categories [15]. An increasing number of HRI studies have used categorical models for classifying emotional expressions such as facial expressions [46] [32] [14] [30], voice [21] [27], body language [34], etc. [42]. However, it is now accepted that the current set of categories are too limited to adopt for real-world emotion recognition of humans [3]. As stated in a recent review, "humans can display a large variety of affective states during social interactions, and affect-detection systems that can identify only a small set of states (e.g., two to five) will not allow a robot to effectively participate in natural bi-directional affective communications with humans [33]."

Dimensional models define emotional expressions as a feature vector in a continuous space. [33]. One of the most recognized dimensional models is the circumplex model of emotion [40]. In this model, two orthogonal dimensions known as pleasure and arousal are used for affect categorization. In another study, Plutchik presented a 3-dimensional model containing eight core emotions, arranged in opposite pairs: sadness and joy, anger and fear, expectation and surprise and an additional dimension as intensity [38]. Another study defined the Pleasure-Arousal-Dominance (PAD) model which consisted of three nearly orthogonal dimensions. Dominance described whether the person feels controlling or controlled [35]. Dimensional models are often harder to understand than categories. For example, [11] states that "categorical labels such as amusement better capture reports of subjective experience than commonly measured affective dimensions."

An example of a hybrid model is the Geneva Emotional Wheel, which combines dimensional and categorical models. This model consists of 20 distinct emotion families. This model also has two dimensions corresponding to valence (horizontal axis) and control (vertical axis) [12]. Although all of these representations are beneficial for improvements and progression in the HRI field, a number of considerations remain to be addressed.

**The benefits of categories**. While dimensional models have been favoured in recent affective computing literature [36] [25], categories allow researchers to clearly distinguish different emotional expressions from each other. [11] suggests that emotional experience is more accurately conceptualized in terms of categories. Furthermore, categories are more likely to be easier to use for the purposes of event-driven robot adaptation to human expressions. The limitation of categorical models is the lack of ability to classify any ambiguous affect that is not included as a category in the model. To overcome this gap, we propose a *data-driven approach* to discover the emotional expressions as they occur in our target domain. Data-driven approaches focus more on data and try to extract relations directly from data [37]. Likewise, we aim to obtain our emotion classes directly from data of humans interacting with a social robot.

**In-the-wild data.** Another limitation that should be addressed is that laboratory conditions have been more widely used up to this time in emotion studies, and the majority of HRI studies use prototypical emotion classes [23]. The most frequent discrete affect states used by robots in HRI settings are disgust, sad, surprise, anger, disgust, fear, sad, happy, surprise, and neutral [41]. We therefore aim to develop a new taxonomy based on the data of humans interacting with a robot in the wild, to increase the number of known classes. To this end, we use a dataset containing recordings of participants interacting with the Pepper[1] robot in a *real world public space*. In addition, we propose the use of emojis as emotion categories. Since emojis are language-agnostic, universally recognized, and used every day in-the-wild, they are an outstanding proxy for emotional expressions [2].

**Dynamic vs. static expressions.** Another limitation is that although some HRI studies use dynamic data [6] [47], many HRI studies performing facial expression recognition use static and frozen in time images of prototypical emotional expressions [31] [9]. These images only show part of the behaviour. A key feature of emotional expressions is their dynamic nature [39]. It seems that using dynamic information is more helpful when expressions are low intensity. In addition, dynamics enable perceivers to observe how expressions change over time. It also helps to differentiate between authentic and fake expressions [26]. Therefore in this study we focus on short video clips ($\leq$ 10s) instead of images to preserve the temporal sequence of emotional expressions.

The aim of this work is therefore to discover the emotional expression categories present during in-the-wild human-robot interactions, focusing on dynamic social signals.

## 2 PRELIMINARY STUDY

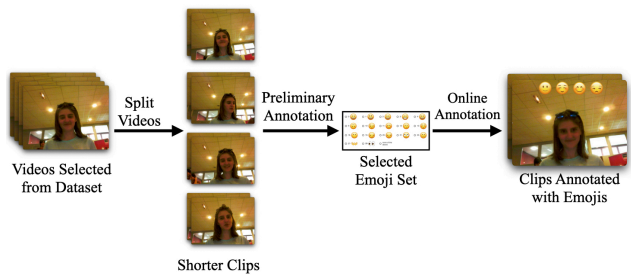

**Figure 1: Proposed data-driven method to discover human expressions in HRI in the wild**

Given an in-the wild dataset that consists of recordings of humans interacting with a social robot, we aim to find a new categorical taxonomy utilizing emoji-based annotations. This process involves three steps. First, we segment and pre-process the raw data streams into short video clips. Second, we conduct a pre-selection phase by asking researchers to label clips by assigning emojis to each clip. Finally, we run a validation phase asking online annotators to annotate clips using the emojis selected in the previous step as a basis.

### 2.1 Dataset

Several studies, for instance [22], [43], [8] and [28] have introduced multimodal human-robot interaction datasets. One of the serious limitations of these studies is that these studies are conducted in the lab. We are most interested in in-the-wild human-robot interaction datasets, such as [7] which consists of 133 interactions which are not publicly available, and [19] [29], two publicly available in-the-wild datasets. While our ultimate goal is to analyze a broad range of datasets, as a first step, we select an in-the-wild, publicly available dataset which focuses on social signals, the User Engagement in HRI (UE-HRI) dataset [4]. This dataset contains 195 recordings (54 among them are publicly available) of participants interacting with the Pepper robot. The robot is placed in a public space and participants are free to start and end the interaction whenever they want. The recording starts automatically when the robot recognizes the presence of a person. If the participant confirms the agreement presented on the robot's tablet, the recorded data is sent to the local server. First, the Pepper robot presents itself and gives instructions. Then, the robot asks questions from participants and talks about their favorite book or movie. The robot records the interaction using all of Pepper's data streams, which are then packaged into the Robot Operating System (ROS[2]) framework's ROSbag[3] file format. The dataset consists of 54 interactions (36 males, 18 females), where 32 are mono-user and 22 are group interactions [4].

### 2.2 Segmentation and Pre-processing

Since the UE-HRI dataset contains raw data streams recorded by the robot packaged into ROSbags, we wrote a script that extracts synchronized front images and turns the image sequence into a video (excluding audio) using ffmpeg [4] [5]. In order to avoid erroneous data, e.g. participants working with the tablet instead of talking to the robot, participants taking video of the robot, etc., we then manually selected 27 of the 54 videos of interactions; in these videos, participants completed the scenarios, meaning they did not leave the robot unexpectedly. The combined length of these 27 videos was 278.7 minutes. In order to ease the final annotations, we performed a semi-automatic segmentation of the videos into clips, each containing, ideally, a single emotional expression. We aimed for clips of less than 10 seconds because upon manual inspection we found that clips longer than 10 seconds frequently contained several emotional expressions. The scene-detection method used for automatically splitting videos into clips was PyScene-Detect [5]. This library has a content-aware detection mode which finds moments where the difference between two subsequent frames exceeds a threshold value and cuts the video at these points. We first used this tool with threshold $T = 30$ which is the default threshold. In the next step, we removed clips with one frame and clips where the face was not visible, resulting in 627 clips. Since there still remained 156 clips longer than 10 seconds, and these long clips often showed more than one social signal, we repeated this process with threshold $T = 20$ resulting in 351 clips, with 69 videos longer than 10s. As a final step we applied threshold $T = 12$ on the remaining 69 long clips. Overall, we acquired 842 total clips of less than 10 seconds, with a total combined length of 77.55 minutes, corresponding to 27% of the 27 selected videos from dataset. The process of choosing optimal threshold values was done manually through trial and error.

---

[1]softbankrobotics.com
[2]http://wiki.ros.org/
[3]http://wiki.ros.org/rosbag
[4]http://www.ffmpeg.org
[5]https://github.com/Breakthrough/PySceneDetect

### 2.3 Emojis as Emotion Expression Classes

In this work, we target emotional expressions as a starting point, but are interested in all social signals used for communication between humans and robots. Instead of relying on traditional emotion models, we use a set of social signals in use every day by people around the world: emojis, otherwise known as smileys. The Unicode Emoji set[6] contains a rich palette of expressions of over 100 face and hand gestures that are frequently updated as new communicative needs are established. In [17], a machine learning system was proposed to convert images of faces into 16 different emojis. Building on this idea, our aim is to discover which of the many other emojis can be used as a representation and annotation tool for dynamic, in-person interactions with robots.

### 2.4 Result of Preliminary Study

The Unicode emoji dataset contains 150 emojis of body and face. In order to decrease cognitive load on the annotators, we performed pre-selection of emojis. As mentioned in Section 2.2, in the first step we manually selected 27 of 54 interactions. In this step, we randomly selected up to 3 clips (after the pre-processing phase, some of the videos had only one remaining clip) from each of these 27 videos in order to cover a broad selection of interactions. The result was 61 clips.

Each clip was annotated independently by two researchers given the full Unicode Emoji set. They selected up to 5 emojis from the given set and specified the confidence of their choice based on a Likert-scale from 1 (not confident) to 5 (very confident). We then aggregated the data into a set of emojis containing all of the emojis both researchers used to annotate videos. The obtained set comprised 29 emojis. We computed the inter-rater agreement between two researchers using Cohen's kappa [10] measure. The agreement was 0.39 which can be interpreted as fair agreement according to [10] [44].

## 3 EXPERIMENT

Annotators were recruited from Amazon Mechanical Turk (MTurk). MTurk is a crowdsourcing website which allows researchers to pay a fee to the MTurk workers to complete Human Intelligence Tasks (HITs). HITs may include surveys or any other work that requires human intelligence. Previous studies have shown that although MTurk samples are not representations of the general population, these samples tend to be more diverse with respect to education and age than college samples [24]. In this experiment, we posted a HIT (Human Intelligence Task) on MTurk containing the 61 clips that researchers had previously annotated. To increase the likelihood of high quality annotations, we set our requirements to a minimum HIT approval rate of ≥ 95% (percentage of completed work that has been approved by other requestors) and having approved HITs ≥ 5000. We required all of our annotators to be located in Canada for ethical board approval. The assignment duration for each task was set to 1 minute and each video clip was annotated by 10 unique annotators. Annotators first provided informed consent about study details and procedure. Then they viewed each of these clips along with the set of 29 emojis obtained from Section 2.4. They were asked to choose one emoji that best matches what the person is expressing to the camera and also specify confidence of their choice based on the Likert-scale from 1 (not confident) to 5 (very confident). They could also write the ID of more than one emoji if they observed multiple expressions, or they could describe the expression if it was not available. Each annotator was paid per HIT such that their per hour rate was equivalent of the minimum wage in their country. The final number of annotations was 610.

---

[6]http://unicode.org/emoji/charts/full-emoji-list.html

## 4 INITIAL RESULTS

The annotators used all of the emojis in the provided emoji set except the waving hand (👋). This means that each of the emojis was perceived in at least one clip by one annotator.

As shown in Figure 4 the results demonstrate that most common emoji is (🙂) which is a slightly smiling face and the second most common emoji is (😐) which is the neutral face. This shows that in most of the dataset's videos, the participants interacting with the robot were slightly smiling or were neutral.

Figure 3 shows the mean of confidence of each emoji. It can be seen in this figure that the 6 emojis that annotators were most confident about were (😄, 😂, 😉, 😮, 🫣, 👀) and 6 emojis annotators were least confident about were (😌, 🙄, 🤔, 😳, 🙂, 😏).

We used the Fleiss' kappa [18] as the statistical measure for inter-rater agreement between more than two raters. This measure is used when data consists of categorical data. The inter-rater agreement for the experiment was 0.123 which is a slight agreement. Our results demonstrated that if we use the higher level categories specified in Figure 2 such as smile, neutral, worried, etc., and combine corresponding emojis into same class, the agreement changes to 0.26 which is fair agreement. This suggests that the lower agreement value in the main experiment was due to a higher number of categories. This shows that annotators in our experiment perceived subcategories of each category differently and the perception of these social signals differ in narrower categories.

### 4.1 An Initial Taxonomy

We devised a hierarchical taxonomy from the initial results obtained, as shown in Figure 2. Our emoji annotation results illustrated that expressions in HRI are not just emotional, therefore, our taxonomy comprises social signals in a broader sense. Researchers discussed until agreement on an initial categorization. As detailed in Figure 2, in the first step, we categorized social signals in the dataset into 6 categories: Smile, Neutral, Worried, Surprised and Other. Inspiration was taken from the taxonomy of the Unicode Emoji dataset. Once the initial categorization had been done, we then sub-categorized social signals into fine-grained sub-categories. From Figure 2 we can see that sub-categories include Duchenne, Polite, Fake and Embarrassed smiles and Confused, Concerned and Skeptical emotional expressions. The Other category includes the social signals that have no clear emotional meaning.

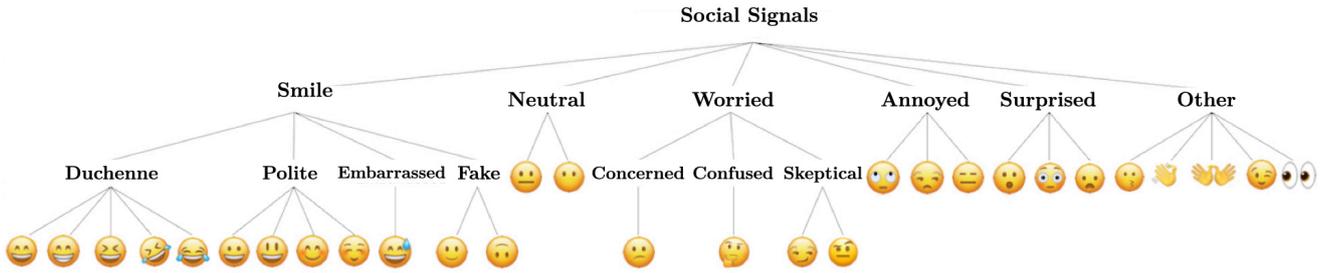

Figure 2: Initial taxonomy of social signals in the UE-HRI dataset

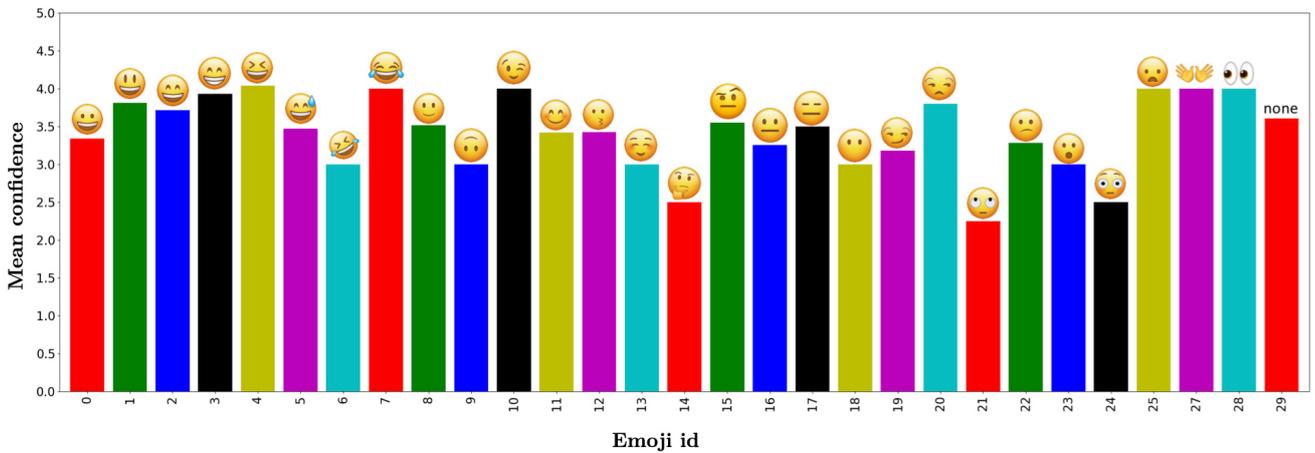

Figure 3: Mean confidence of each emoji obtained by annotations in Mturk

Figure 4: Frequency of emojis obtained by annotations in MTurk

## 5 DISCUSSION AND FUTURE WORK

We have created an initial taxonomy of social signals and emotional expressions in real-world human-robot interactions utilizing emojis as expression classes. The findings of this study suggest that there are additional emotion categories present in real-world social settings other than the existing prototypical categories. In addition to neutral and smiling expressions, we found a relatively large number of skeptical expressions. One limitation of our study is that the people interacting with the robot in the UE-HRI dataset were primarily students located in France and speaking in French. Future work may take into account culture and language when considering expressions and annotation, and applying our method to other available in-the-wild datasets. Another limitation is the matching of textual descriptions of emotions to the emojis, which remains to be developed. To further our research, we also plan to explore the 'no emoji' or 'other' cases, and analyze the full dataset in order to enhance our taxonomy. Future studies may also study the varied contexts (e.g. group vs solo) of HRI. In summary, we have created an initial taxonomy of emotional expressions by using emojis as emotion models in real-world human-robot interactions.

## ACKNOWLEDGEMENT

This work was supported by the Natural Sciences and Engineering Research Council of Canada Discovery Grants program.